# Noise-Aware Training of Layout-Aware Language Models

**Ritesh Sarkhel**
The Ohio State University

**Xiaoqi Ren**
Google

**Lauro Beltrao Costa**
Google

**Guolong Su**
Google

**Vincent Perot**
Google

**Yanan Xie**
Google

**Emmanouil Koukoumidis**
Google

**Arnab Nandi**
The Ohio State University

## Abstract

A visually rich document (VRD) utilizes visual features along with linguistic cues to disseminate information. Training a custom extractor that identifies named entities from a document requires a large number of instances of the target document type annotated at textual and visual modalities. This is an expensive bottleneck in enterprise scenarios, where we want to train custom extractors for thousands of different document types in a *scalable way*. Pre-training an extractor model on unlabeled instances of the target document type, followed by a fine-tuning step on human-labeled instances does not work in these scenarios, as it surpasses the maximum allowable training time allocated for the extractor. We address this scenario by proposing a **N**oise-**A**ware **T**raining method or `NAT` in this paper. Instead of acquiring expensive human-labeled documents, `NAT` utilizes weakly labeled documents to train an extractor in a *scalable way*. To avoid degradation in the model's quality due to noisy, weakly labeled samples, `NAT` estimates the confidence of each training sample and incorporates it as uncertainty measure during training. We train multiple state-of-the-art extractor models using `NAT`. Experiments on a number of publicly available and in-house datasets show that `NAT`-trained models are not only *robust in performance* – it outperforms a transfer-learning baseline by up to 6% in terms of macro-F1 score, but it is also more *label-efficient* – it reduces the amount of human-effort required to obtain comparable performance by up to 73%.

## 1 Introduction

A visually rich document (VRD) refers to a physical or digital document that utilizes explicit (e.g. *font size, font color*) and implicit (e.g. *relative positioning, whitespace alignment*) visual features, along with linguistic cues to disseminate information effectively. VRDs such as invoices, tax forms, utility bills, infographic posters, insurance quotes

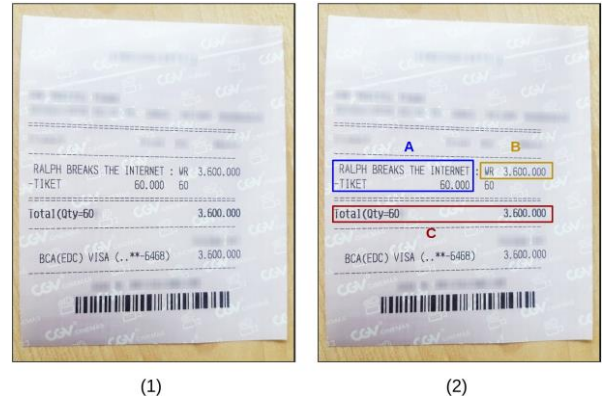

Figure 1: The figure on the left shows a receipt from the CORD dataset (Park et al., 2019). The figure on the right shows some of the key entities appearing on this document, e.g. A shows the name of the purchased item, B shows its price

are common in day-to-day business workflows. Take the case of invoices for instance (see Fig. 1). Large enterprises process thousands of invoices every week (iPayables, 2020). Invoices from different vendors present similar kind of information in different layouts and formats. As this information cannot be retrieved from other sources easily, automating information extraction (IE) from such documents reduces the amount of human-effort it would require to parse these documents manually.

Implementing a generalizable solution for this task is challenging. Off-the-shelf approaches that work well on unstructured text do not apply as it requires understanding both textual and visual properties of the document (see Fig. 1). Moreover, as a significant portion of VRDs are available in PDF or image format, spatial relationships between different elements cannot be explicitly defined using a markup language. This makes techniques such as wrapper induction (Chang et al., 2006) that work well on HTML documents hard to translate for this extraction task.

Contemporary researchers have proposed a number of different solutions for this task. Sarkhel and Nandi (2019) proposed a segmentation algorithm that partitions a document into semantically similar visual areas and then searches for a set of syntactic/semantic patterns within each area. Davis et al. (2019) developed a fully convolutional network to infer the semantic structure of a document. Liu et al. (2019b) developed a graph convolution network to encode pairwise relationship between different visual elements. Sarkhel and Nandi (2021) and Wu et al. (2018) developed multimodal bi-LSTM networks to jointly encode the visual and textual context of a token. Katti et al. (2018) proposed a convolutional approach by taking the positional information of each individual character in the document into account. The aforementioned methods work in a supervised way. To train a custom extractor, we require a large number of human-labeled samples from the target document type. Labeling a document with sufficiently high accuracy, however, takes significant human-effort. Hwang et al. (2021a) estimates the cost of manually labeling a VRD to be $\approx 0.08$ person-hour. This is an expensive bottleneck in enterprise scenarios where we want to train custom extractors for thousands of different document types in a *scalable way*. A plausible way to address this is to pre-train the extractor model on unlabeled instances of the target document type and then fine-tune it on human-labeled samples. In fact, there has been a recent surge in research interest in pre-training strategies for VRDs. Recent developments in this direction e.g. Juno (Sarkhel and Nandi, 2023), LayoutLMV2 (Xu et al., 2020), DocFormer (Appalaraju et al., 2021), BROS (Hong et al., 2021), SPADE (Hwang et al., 2021b) UniDoc (Gu et al., 2021), SelfDoc (Li et al., 2021), LiLT (Wang et al., 2022), FormNet (Lee et al., 2022), Least (Sarkhel et al., 2023) all utilize a number of unsupervised multimodal objectives to pre-train an extractor model and then fine-tune it on human-labeled samples pertaining to the downstream extraction task. We describe the pre-training strategy adopted by two state-of-the-art extractors in Section 3. This approach, however, is not viable in scenarios where the maximum allowable training time allocated for an extractor has a strict upper bound, as the total time needed for pre-training often surpasses this bound[1]. This is the scenario we address in this paper. The key research question we address in this paper is as follows – *how to train a robust extractor with limited number of human-labeled samples within a bounded time?*

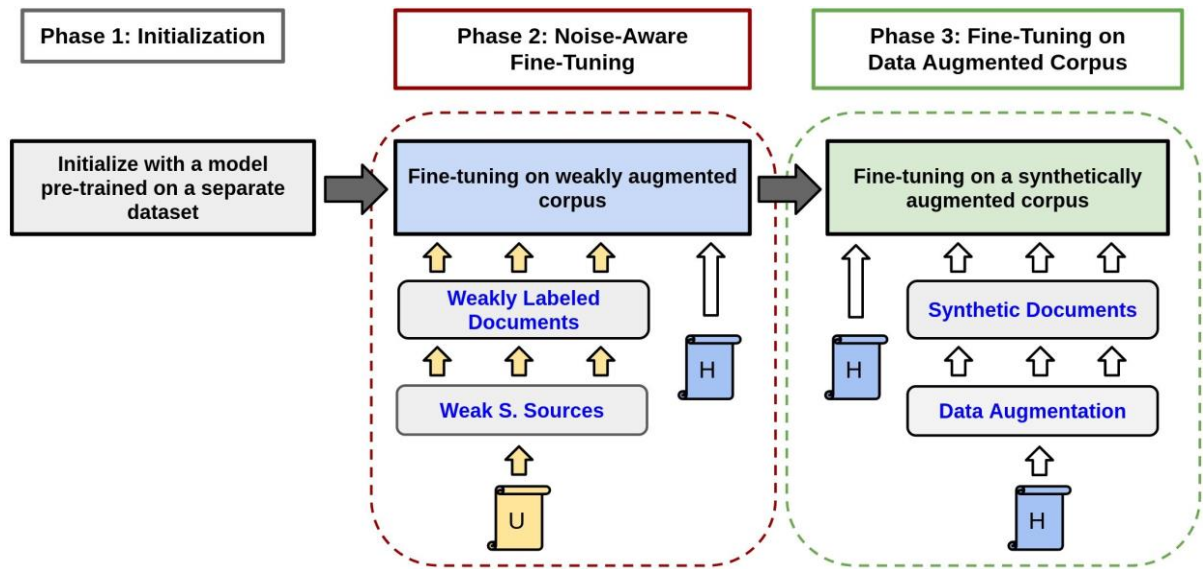

Figure 2: An overview of the continual training framework adopted by NAT. After initializing the extractor $E$ with a pre-trained model, it assigns labels to unlabeled documents ($U$) using a number of weak supervision sources, supplements them with a limited number of human-labeled documents ($H$) and fine-tunes the model on this weakly augmented corpus ($H \cup U$). Finally, it generates synthetic documents ($S$) and fine-tunes the model on this synthetically augmented corpus ($H \cup S$).

To this end, we propose a semi-supervised continual training method. It works in three phases.

---
[1]Pre-training the LayoutLMV2 model (Xu et al., 2020) on the IIT-CDIP (Lewis et al., 2006) dataset on a Titan XP GPU took us $\sim$ 30 hours

In `Phase I`, we initialize the extractor following a L1-transfer (Pan and Yang, 2010) of weights from a model pre-trained on a separate dataset offline (Section 3). In `Phase II`, we assign weak labels to some unlabeled documents ($U$) of the target type (e.g. invoices), supplement them with a few human-labeled documents ($H$) and fine-tune the pre-trained model on this weakly augmented corpus $H \cup U$. To prevent gradual drifts in the model's quality due to training on noisy training samples (Ruder and Plank, 2018), we introduce a *noise-aware training* scheme at this stage (Section 4). Finally in `Phase III`, we generate synthetic documents ($S$) using a heuristics-based data augmentation strategy, supplement them with human-labeled documents ($H$) and fine-tune the model obtained from `Phase II` on this synthetically augmented corpus $H \cup S$ (Section 5). We implement all three phases within a single framework. We refer to this framework as `NAT`. Contrary to existing works on cost-effective information extraction that rely on pre-existing dictionaries (Hwang et al., 2021a; Wang et al., 2021) as a source of supervision or rely on document layout similarity (Cheng et al., 2020; Yao et al., 2021), `NAT` takes a model-based approach for weak supervision, allowing multiple models to act as independent sources of weak supervision within its framework without extensive human-effort.

We train two state-of-the-art extractor models using `NAT` on a number of VRD datasets for separate extraction tasks. The key takeaways from our experiments are as follows. *First*, `NAT` is generalizable to multiple existing models. *Second*, `NAT` is *label-efficient* – a `NAT`-trained model reduces human-labeling effort to obtain comparable performance in a supervised setting by up to 73%. *Third*, a `NAT`-trained model is *robust* in terms of extraction performance – it outperforms a transfer learning-based baseline by up to 6% in macro-F1 score. *Fourth*, `NAT`-training is fast – we can train a custom extractor model using `NAT` within ∼1.5 hours on a Titan XP GPU. This is within the maximum allowable training time in the enterprise scenarios mentioned before. We present our results in greater details in Section 4. We formalize the research problem addressed in this paper next.

## 2 Problem Definition

Given a set of named entities $N$ and VRDs of a target type $T$, our goal is to train an extractor $E$ that identifies a set of text-spans from each document for every $n_i \in N$ with three overarching goals.

i $E$ is *robust* in terms of end-to-end extraction performance.

ii $E$ can be trained with a limited amount of human-labeled documents.

iii The maximum allowable training time for $E$ is *constrained by an upper bound $t_{max}$*.

We develop a semi-supervised, continual training method, called `NAT` to address this. It initializes the extractor model $E$ with pre-trained weights from a separate dataset.

## 3 Phase I: Pre-training VRD Extractors

We provide a brief introduction of the pre-training strategy adopted by two state-of-the-art extractor models used in our experiments – LayoutLMV2 and FormNet next.

### 3.1 LayoutLMV2

LayoutLMV2 (Xu et al., 2020) adopts a large-scale Transformer-based architecture for VRD related downstream tasks. The model has a total of 426M learnable parameters. It encodes relative positions as well as pairwise relationships between different elements in the document by using a two-stream (one each for encoding visual and textual features), Transformer-based encoder with spatially-aware self-attention. For pre-training, it uses three unsupervised task objectives. They are as follows. The *masked visual-language modeling* task randomly masks some of the tokens in the document and asks the model to recover them while keeping the layout information unchanged. The *text-image alignment task* randomly selects some tokens, covering their corresponding image regions and asks the model to predict if a token has been covered or not. Finally, the *text-image matching task* feeds the output representation of the `[CLS]` token into a classifier which predicts whether an image and text are from the same document page. Once pre-trained, task-specific heads are added over the text part of model output (similar to Devlin et al. (2019)) and the model is fine-tuned on human-labeled samples pertinent to the extraction task. For more background on this model, we point the readers to the original work by Xu et al. (2020).

## 3.2 FormNet

FormNet (Lee et al., 2022) is a ETC-Transformer (Ainslie et al., 2020) based model. It is a slightly smaller model than LayoutLMV2 with 345M learnable parameters. Given a document, this model uses a multilingual BERT-tokenizer to tokenize the OCR output. These tokens and their 2D coordinates are then fed into a Graph Convolutional Network (GCN) which constructs super-tokens by merging the neighboring tokens. These super-tokens are fed into a ETC-Transformer based encoder. The advantage of using a ETC-Transformer-based encoder is that it is able to encode long sequences in an efficient way as it replaces the standard attention used by a Transformer-based encoder (Vaswani et al., 2017) with quadratic complexity with a sparse attention mechanism. To address the issue of imperfect serialization in OCR output, it constructs a document graph by connecting the neighbouring tokens to construct a super-token. Edges of this graph are based on an inductive bias that ensures that tokens belonging to the same entity type are connected with high probability. FormNet is pre-trained using the *masked language model* objective similar to Appalaraju et al. (2021), which randomly masks a set of tokens in the document and then asks the model to reconstruct these tokens. To perform entity extraction, it decodes the output representations of each token using the Viterbi algorithm and tags each token following the BIOES entity tagging scheme.

We use the official implementation and pre-trained weights (Xu et al., 2021) of LayoutLMV2$_{BASE}$ released by Xu et al. for our experiments. In case of FormNet, we use a simplified version of the model shared by the authors. In this version, instead of constructing the super-tokens, we feed each token from the OCR'ed output directly to the ETC-Transformer-based encoder. For both models, we follow the hyperparameter settings recommended by their respective authors.

## 4 Phase II: Noise-Aware Fine-Tuning on A Weakly Augmented Corpus

To train an extractor model $E$ in a scalable way, we rely on weak supervision to automatically infer training samples from unlabeled documents. Each training sample corresponds to a text-span appearing in the document and corresponds to an entity type defined for the target document type. We supplement these weak labels with a limited number of human-labeled samples and fine-tune the extractor model on the resulting corpus in Phase II of the training workflow.

### 4.1 Model-based Weak Supervision

Prior works on cost-effective IE from VRDs such as Hwang et al. (2021a) and Wang et al. (2021) have utilized a dictionary of named entities to automatically infer training samples via fuzzy string matching. In our experience, a large enough corpus like this is not always available for a target document type. Researchers like Wu et al. (2018) and Sarkhel and Nandi (2021) have utilized heuristics-based functions to assign weak labels to matching phrases in unlabeled documents. Designing these functions that have sufficiently high coverage, however, requires significant domain expertise. Cheng et al. (2020) and Yao et al. (2021) utilized layout homogeneity to limit the number of training samples required to train an extractor model. This constraint is too restrictive to be generalizable for all document types. We use multiple model-based extractors trained on limited human-labeled samples to infer weak labels from unlabeled documents in our framework.

We use two model-based extractors as our weak supervision source. The first weak supervision source is FormNet (Lee et al., 2022). We pre-train this model on the same unsupervised objectives described in Section 3.2 on the same dataset used to initialize the extractor model $E$, and then fine-tune it on the limited number of human-labeled samples present in our training corpus. Our second weak supervision source is a multimodal bi-LSTM network with attention, similar to the model developed by Sarkhel and Nandi (2021). Along with contextual semantic features, it uses a number of visual features to encode each token. We train this model for an entity extraction task on the human-labeled samples in our training corpus. Once training terminates, we feed unlabeled documents of the target type to the fine-tuned model which infers weak labels from each document. Using models as weak supervision sources obviates the necessary of a comprehensive dictionary like Hwang et al. (2021a), domain expertise to design high-quality labeling function like Wu et al. (2018), or assume layout homogeneity like Yao et al. (2021), instead we focus on the training iterations of each model.

One of the challenges of fine-tuning a model on weakly labeled samples is that labeling errors introduced due to the inductive biases of the supervision source cascades into the model during training iterations. NAT takes a three-prong approach to address the challenges posed by noisy training samples. They are as follows.

### 4.2 Sample Re-Weighting

NAT estimates the confidence of each training sample and uses it to re-weight that training sample during loss computation. Higher weights are assigned to more confident samples and smaller weights are assigned to less confident samples. If a training sample is human-labeled, we assign it a weight equal to 1.0. Otherwise, we assign it a weight $0 < c < 1$ equal to the softmax probability of its label inferred by the corresponding weak supervision source.

### 4.3 Weight Thresholding

To reduce the amount of noisy training samples in the resulting corpus, NAT performs *weight thresholding* on the weakly labeled samples. This entails discarding all training samples with a sample-weight less than a predefined threshold C from the training corpus. The rest of the samples are used in the fine-tuning step. Value of the threshold depends on the weak supervision source.

### 4.4 Noise-Aware Loss

To fine-tune the extractor model $E$ on the resulting corpus, we introduce a *noise-aware loss* function. It uses the sample-weights to adaptively boost or dampen the signal from each training sample based on its estimated label accuracy. More specifically, it boosts signals from samples with higher weights and dampens signals from samples with lower weights. We define the loss function as follows.

$$L_{na}(y, y^{'}) = c \, L(y, y^{'}) + \lambda \, L_0 \qquad (1)$$

In Eq. 1, $L_{na}$ represents the noise-aware loss computed for a training sample with a softmax label $y^{'}$ and groundtruth label $y$, $c$ denotes the sample-weight, L denotes a cross-entropy-based loss, and $L_0$ denotes a regularizer term. $\lambda$ is a hyperparameter. $L_0$ represents the *unlikelihood* of $y$ being a groundtruth label of the sample. We define it as the cross-entropy-based loss computed by the *opposite model* with respect to the extractor model $E$. We define the opposite model of $E$ by following Tizhoosh (2005). It is a model with an identical architecture as $E$ with weights defined as follows.

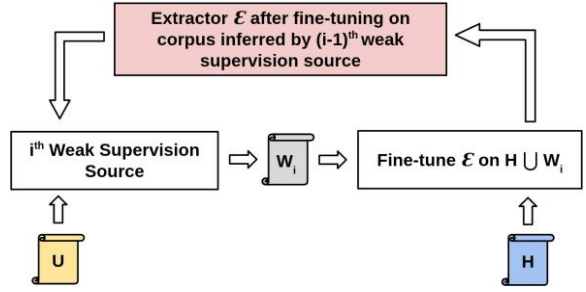

Figure 3: The pre-trained extractor model $E$ is sequentially fine-tuned on a weakly augmented corpus ($H \cup W_i$) inferred by the $i^{th}$ weak supervision source from unlabeled documents ($U$) of the target type

$$p^{'}_{ij}(t) = p^{max}_j(t) + p^{min}_j(t) - p_{ij}(t) \qquad (2)$$

In Eq. 2, $p_{ij}(t)$ represents the $i^{th}$ weight of the $j^{th}$ layer of the extractor $E$ at $t^{th}$ training epoch, $p^{'}_{ij}(t)$ denotes the corresponding weight in the opposite model; $p^{max}_j(t)$ & $p^{min}_j(t)$ represent the maximum and minimum weight of $E$ at $j^{th}$ layer.

### 4.5 Sequential Fine-Tuning

We fine-tune the extractor model $E$ obtain from Phase I on a weakly augmented corpus that contains both weakly labeled and human-labeled ($H$) training samples. In case of multiple weak supervision sources, we sequentially fine-tune the model on weakly augmented corpus inferred by each weak supervision source. For example, if we denote the weakly labeled corpus inferred by our two weak supervision sources as $W_1$ and $W_2$, we fine-tune the extractor model $E$ on the weakly augmented corpus $H \cup W_1$ first. Once training terminates, we fine-tune it on $H \cup W_2$.

## 5 Phase III: Fine-Tuning on Synthetically Augmented Corpus

We fine-tune the model from Phase II on a synthetically augmented corpus ($H \cup S$) using a cross-entropy based loss. We adopt a rule-based data augmentation strategy to generate synthetic documents ($S$) from the human-labeled corpus ($H$). Given a set of human-labeled documents $H$ and a data augmentation rule-set $R$, we apply each augmentation rule $r \in R$ pertinent to the target document type to every human-labeled document $H_i \in H$. Formally, an augmentation rule $r$ takes a human-labeled document $H_i$ as input and outputs a synthetic document $S_{r,i}$. We make five separate passes of the rule-set $R$ over the human-labeled corpus $H$ to construct the corpus of synthetic documents $S$.

## 5.1 Synthetic Data Augmentation

We use four different types of rules to generate synthetic documents. They are as follows.

### 5.1.1 Synonym Substitution

Key phrases used to describe various fields of interest may vary a lot in documents from heterogeneous sources. For example, any of the terms `Total`, `Tot.`, `Amount`, or `Total Amount` may refer to the total billed amount in an invoice. Given a human-labeled document $H_i$, a synonym substitution rule replaces a set of common key-phrases in the document with a phrase randomly sampled from its corresponding set of synonyms..

### 5.1.2 Format Substitution

Various fields of interest can represent the same information in different formats. For example, the date of purchase can appear as `11/19/90`, `11-19-90`, or `19`th `Nov, 90` in an invoice. Given a human-labeled document $H_i$, a format substitution rule stochastically replaces a field of interest with a phrase randomly sampled from one of its many alternative formats with equal probability..

### 5.1.3 Coordinate Transformation

Beyond textual information, the position of a field of interest also contains valuable information about an entity type. For example, an instance of the entity `purchase_date` often appears at the beginning of an invoice, whereas an instance of the entity `total_billed_amount` appears at the latter half. A coordinate transformation rule shifts the coordinates of a token by a random fraction of the document's dimension horizontally and/or vertically.

### 5.1.4 Bounding-Box Expansion

The font size used for a field of interest varies depending on the source of the document. For example, the entity `vendor_name` appears in larger font sizes in an invoice. A bounding-box expansion rule expands the bounding-box of some fields of interest in both vertical and horizontal directions by a random fraction.

## 6 Experiments

We seek to answer four key questions in our experiments. *First*, we answer how *robust* a `NAT`-trained extractor model is by evaluating its end-to-end extraction performance on four different datasets. *Second*, we answer how *label-efficient* `NAT` is by comparing the number of human-labeled samples required to train the same extractor model in a supervised setting to obtain a comparable performance. *Third*, we answer how does a `NAT`-trained extractor model perform against existing approaches by comparing it against a number of strong baselines trained under similar settings. *Fourth*, we measure the individual contribution of each component in our workflow on end-to-end extraction performance by performing an ablation study.

## 6.1 Datasets

We evaluate the performance of a `NAT`-trained extractor model on two publicly available and two in-house datasets that we collected.

**CORD.** CORD (Park et al., 2019) or the *Consolidated Receipt Dataset for Post-OCR Parsing* contains thousands of Indonesian receipts from different sources. It is a publicly available dataset consisting of 1000 documents annotated at word-level for 30 different entity types, e.g. `store_name`, `discount` etc.

**FUNSD.** FUNSD (Jaume et al., 2019) or the *Form Understanding in Noisy Scanned Documents Dataset* is a publicly available dataset consisting of 199 scanned forms in English annotated at word-level for 4 different entity types, e.g. `header`, `question`, `answer`, & `other`.

| Dataset | #Train | #Test | #Unlabeled |
|---|---|---|---|
| CORD | 500 | 100 | 100 |
| FUNSD | 49 | 50 | 100 |
| Utility | 240 | 100 | 140 |
| French Invoice | 500 | 5,000 | 2,700 |

Table 1: Number of documents in the training, test, and unlabeled corpus for each dataset

**In-house datasets.** We collect two datasets for our experiments. The *French Invoice Dataset* contains ~ 5,500 invoices in French from different vendors annotated at word-level with X different entity types such as `total_amount`, `invoice_date` etc. The *Utility Dataset* contains ~ 700 utility bills in English annotated at word-level with Y different entity types e.g. `total_due`, `customer_name` etc.

## 6.2 Experiment Design

**Pre-training.** We use two state-of-the-art models as the backbone in our experiments. We initialize the LayoutLMV2 model following a L1-transfer of weights from a model pre-trained on the IIT-CDIP dataset (Lewis et al., 2006). For extraction tasks on the two publicly available datasets, we pre-train FormNet on the IIT-CDIP dataset. This dataset contains ∼ 6.9M scanned document images from the "tobacco collection" created for the TREC Legal Task. For extraction tasks on the in-house datasets, we pre-train the FormNet model on a dataset of English invoices that we collected. This dataset contains Z invoices from different vendors.

**Dataset split.** We describe the size of human-labeled documents in the training and test corpus for each dataset in Table 1. The last column in this table represents the number of unlabeled documents used to infer weak labels in our training workflow for each dataset.

**Evaluation.** We evaluate the end-to-end extraction performance of each model by reporting its macro-F1 score over all entity types defined for a dataset. For every dataset, we repeat our experiment 9 times and report the average and standard deviation of the macro-F1 score of each competing model. To measure the label-efficiency of a NAT-trained model, we compute the number of human-labeled samples NAT saves, by comparing the number of human-labeled samples in its training corpus with the number of human-labeled training samples it would take to obtain a comparable performance from the same model trained without NAT.

## 6.3 Experimental Results

**End-to-end performance.** We report the macro-F1 score of both LayoutLMV2 and FormNet-based models trained using NAT on all datasets in Table 2. We observe that a NAT-trained FormNet model outperforms LayoutLMV2 on both of the public datasets. This is due to the inherent difference in the learning capabilities of these two models trained under same settings. This is congruent to findings of Lee et al. (2022). Increasing the number of human-labeled documents improves end-to-end extraction performance on all datasets (see Fig. 4). Small standard deviation in the extraction performance repeated for 9 trials (see last column of Table 2) suggests that an extractor trained using NAT is stable in its extraction performance.

| Model | Dataset | Avg. F1 | Std. |
|---|---|---|---|
| FormNet | CORD | 94.80 | 0.003 |
| | FUNSD | 76.85 | 0.007 |
| | Utility | 85.70 | 0.006 |
| | French Invoice | 59.60 | 0.013 |
| LayoutLMV2 | CORD | 91.58 | 0.005 |
| | FUNSD | 75.32 | 0.005 |

Table 2: Number of human-labeled and unlabeled documents in each dataset

**Label efficiency.** We measure the label efficiency of a NAT-trained model by computing the percentage of additional human-labeled samples needed it would be required to fine-tune the same pre-trained model used to initialize our training workflow and obtain a comparable extraction performance as shown in Table 2. The final column in Table 3 shows the percentage of human-labeling effort saved for each dataset. We observe a significant reduction in human-labeling effort for all datasets.

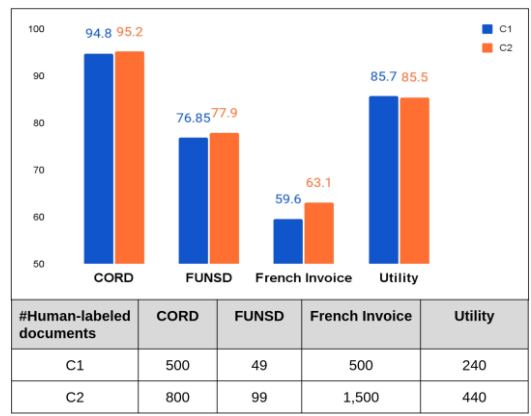

| #Human-labeled documents | CORD | FUNSD | French Invoice | Utility |
|---|---|---|---|---|
| C1 | 500 | 49 | 500 | 240 |
| C2 | 800 | 99 | 1,500 | 440 |

Figure 4: Average macro-F1 score of a NAT-trained FormNet model on varying sized human-labeled corpus

| Dataset | #Human-labeled | #Weakly labeled | ∼ Saved (%) |
|---|---|---|---|
| CORD | 500 | 100 | 40 |
| FUNSD | 49 | 100 | 35 |
| Utility | 240 | 140 | 66 |
| French Invoice | 500 | 2,700 | 73 |

Table 3: Number of documents in the training, test, and unlabeled corpus of each dataset

**Training time.** Our results show that a NAT-trained extractor model is robust in terms of extraction performance as well as efficient in reducing human-effort in its end-to-end training workflow. To be scalable to different document types, however, a custom extractor also has to be trained within a maximum allowable time. Training a custom extractor using NAT on a Titan

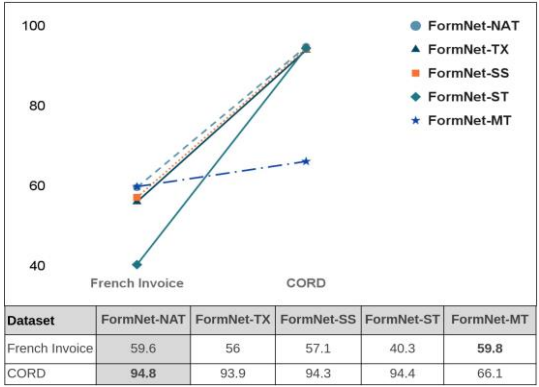

Figure 5: Average macro-F1 score of all competing models on CORD and French Invoice dataset

| Dataset | FormNet-NAT | FormNet-TX | FormNet-SS | FormNet-ST | FormNet-MT |
|---|---|---|---|---|---|
| French Invoice | 59.6 | 56 | 57.1 | 40.3 | **59.8** |
| CORD | **94.8** | 93.9 | 94.3 | 94.4 | 66.1 |

XP GPU took us a maximum of ∼1.5 hours. This is within the maximum allowable training time in the enterprise scenario we are interested in.

**Comparison against baselines.** We compare the extraction performance of a `NAT`-trained FormNet model (highlighted column in Fig. 5) against a number of label-efficient baselines on a publicly available dataset – the CORD dataset and an in-house dataset – the French Invoice dataset. In Fig. 5, `FormNet-TX` represents a transfer-learning-based baseline similar to Lee et al. (2022). After initializing the extractor model with pre-trained weights (Section 3.2), we fine-tune this model only on human-labeled samples. `FormNet-SS` represents a semi-supervised baseline that utilizes the `FormNet-TX` model to infer weak labels from unlabeled documents of the target type. The pre-trained model is fine-tuned on this weakly augmented corpus using cross-entropy-based loss, followed by a fine-tuning step on the synthetically augmented corpus (Section 5).

`FormNet-ST` follows a self-training approach similar to Sara et al. (2022). It initializes the teacher model with weights of the `FormNet-TX` model. The teacher model then infers weak labels & initializes the student model. The student model is fine-tuned on a weakly augmented corpus containing these weak labels & human-labeled samples. Once training terminates, the student's weights are used to initialize the teacher model. These iterative knowledge exchange is repeated until termination. Following the recent success Clark et al. (2018); Liu et al. (2019a) of multi-task training in label-scarce scenarios, `FormNet-MT` introduces auxiliary task-heads along with the primary entity extraction task and fine-tunes the pre-trained extractor model on the human-labeled samples in the training corpus in a multi-task setting. We describe the meta-learning algorithm used to infer the auxiliary tasks for this model in Appendix A. We train all competing models on the same human-labeled/unlabeled documents. We observe that the `NAT`-trained FormNet model outperforms 3 out of 4 of these baselines and performs comparatively to the larger `FormNet-MT` model on the French Invoice dataset.

| Scenario | Dataset | F1(%) | ΔF1↓(%) |
|---|---|---|---|
| NA training ✗ | CORD | 94.30 | 0.50 |
| | French Invoice | 57.10 | 2.50 |
| Synthetic docs ✗ | CORD | 94.50 | 0.30 |
| | French Invoice | 58.60 | 1.0 |
| Weak labels ✗ | CORD | 93.90 | 0.90 |
| | French Invoice | 56.50 | 3.10 |

Table 4: Results from the ablation study

**Ablation study.** We measure the individual contribution of three different components in our workflow on end-to-end extraction performance by performing an ablation study in Table 4. In the first scenario, we *remove the noise-aware training scheme* (Section 4.2) from our workflow, resulting in a 2.50% drop in average macro-F1 score in the French Invoice dataset. We observe that the drop in performance correlates to the size of the weakly labeled corpus. In the second scenario, we remove *fine-tuning on the synthetically augmented corpus* from our workflow. This worsens extraction performance for both of our datasets. In the third and final scenario, we remove the fine-tuning step on the weakly augmented corpus and fine-tune the pre-trained model directly on the synthetically augmented corpus. Results show that using weak labels in our workflow has serious implications on the extraction performance of a `NAT`-trained model. For the French Invoice dataset, this results in a significant 3.10% decrease in macro-F1 score.

## 7 Limitations

One of the challenges of implementing a `NAT`-training workflow is to identify supervision sources that can construct a large enough corpus of weakly labeled documents without requiring significant human-effort or domain expertise. If high quality supervision sources can not be identified, simply increasing the number of the unlabeled documents do not lead to improved extraction performance. We study the impact of the number of unlabeled documents in `NAT`'s workflow in Appendix A.

## 8 Conclusion

Training a custom extractor for thousands of different document types in an enterprise scenario requires a training workflow that is *fast, requires minimal human-effort* while ensuring *robust extraction performance*. `NAT` *addresses all these within a single framework*. It utilizes a noise-aware training scheme to train a custom extractor on weak labels inferred by multiple supervision sources. Contrary to prior works, `NAT` neither requires a pre-existing dictionary nor significant domain-expertise to generate weak labels. Exhaustive experiments on four different datasets for separate extraction tasks show that `NAT`-trained models ensure robust extraction performance while reducing the amount of human-labeling effort to obtain comparable performance in a supervised setting by up to 73%.